%% file: RVDRL.tex
\documentclass[letterpaper]{article}
\usepackage{bm}
\usepackage{url}
\usepackage{tikz}
\usepackage{aaai}
\usepackage{times}
\usepackage{xcolor}
\usepackage{xspace}
\usepackage{subfig}
\usepackage{helvet}
\usepackage{amssymb}
\usepackage{courier}
\usepackage{caption}
\usepackage{mdwlist}
\usepackage{diagbox}
\usepackage{amsmath}
\usepackage{amsfonts}
\usepackage{stfloats}
\usepackage{pgfplots}
\usepackage{tabularx}
\usepackage{multirow}
\usepackage{graphicx}
\usepackage{booktabs}
\usepackage{microtype}
\usepackage{blindtext}
\usepackage{graphicx}
\usepackage{subfig}
\usepackage{hyperref}
\usepackage{algpseudocode,algorithm}

\pgfplotsset{compat=newest}

\usetikzlibrary{backgrounds}
\usetikzlibrary{calc}
\usetikzlibrary{external}
\usetikzlibrary{fit}
\usetikzlibrary{positioning}
\usetikzlibrary{shapes}
\tikzexternaldisable

\input{header}

\begin{document}

\title{Stroke-based Character Reconstruction}
\author{Zhewei Huang\textsuperscript{1,2},
  Wen Heng\textsuperscript{1},
  Yuanzheng Tao\textsuperscript{1,2},
  Shuchang Zhou\textsuperscript{1}
  \\ %\footnotemark[1]
\textsuperscript{1}Megvii Inc(Face++)
\textsuperscript{2}Peking University\\
{hzwer@pku.edu.cn, hengwen@megvii.com, memphis@pku.edu.cn, zsc@megvii.com}
}
\maketitle
\begin{abstract}
\begin{quote}
Background elimination for noisy character images or character images from real scene is still a challenging problem, due to the bewildering backgrounds, uneven illumination, low resolution and different distortions. We propose a stroke-based character reconstruction(SCR) method that use a weighted quadratic Bezier curve(WQBC) to represent strokes of a character. Only training on our synthetic data, our stroke extractor can achieve excellent reconstruction effect in real scenes. Meanwhile. It can also help achieve great ability in defending adversarial attacks of character recognizers. 
\end{quote}
\end{abstract}
\input{intro}
\input{related}
\input{env}
\input{exper}
\input{discussion}
\bibliographystyle{aaai}
\bibliography{RVDRL}
\input{appendix}
\end{document}

%% file: header.tex
\makeatletter
\newcommand*{\etc}{%
    \@ifnextchar{.}%
        {etc}%
        {etc.\@\xspace}%
}
\makeatother

\makeatletter
\newcommand{\todo}[1][]{\@latex@warning{TODO #1}}
\makeatother

\newlength\figureheight
\newlength\figurewidth

\newif\ifshowplots
\showplotstrue % comment out to hide answers

\definecolor{dmyellow}{HTML}{F8E46C}
\definecolor{dmlightblue}{HTML}{7DC3D2}
\definecolor{dmred}{HTML}{EF5A58}
\definecolor{dmpurple}{HTML}{DA81B3}
\definecolor{dmgreen}{HTML}{BCD26F}

\usepackage{tikz}
\usetikzlibrary{shapes, arrows, fit}
\usetikzlibrary{positioning,shadows}
\usetikzlibrary{decorations.pathreplacing}
\definecolor{inodefill}{rgb} {0.00,0.53,0.88}
\definecolor{inodedraw}{rgb} {1.00,1.00,1.00}
\definecolor{pnodefill}{rgb} {1.00,1.00,1.00}
\definecolor{onodedraw}{rgb} {0.00,0.00,0.00}
\definecolor{onodefill}{rgb} {1.00,1.00,1.00}
\definecolor{pnodedraw}{rgb} {0.00,0.00,0.00}
\definecolor{mnodefill}{rgb} {0.00,0.53,0.88}

%% file: intro.tex
\section{1.Introduction}
\label{sec:introduction}

Deep neural networks(DNN)\cite{krizhevsky2012imagenet} have  already shown great ability in various applications in computer vision, including image classification, object detection, scene understanding etc. Some of the most influential jobs are VGGNet\cite{simonyan2014very}, GoogLeNet\cite{szegedy2015going} and residual connections\cite{he2015delving}.

Character recognition is the task that converts the printed or hand-written text into machine-encoded text. This task is usually converted into a multi-category classification task, i.e., one character corresponds to one category. So in recent years, many character recognition models based on DNN have been produced. Since most character recognition methods directly predict the labels based on the character images, no more structure information about characters is considered. If we want to extract the text information of the picture, eliminate the background of the picture, or reconstruct the blurred document, then only the classification information is not enough.

Character recognition of natural images is still a challenging problem to machines. It is because it's tough for machines to extract adequate information while the characters are with a bewildering background, uneven illumination, low resolution and different distortions. It's also shown that the recognition of hand-written digits is easily fooled by adding some noise to the character images\cite{madry2017towards}, which is called the adversarial attack. One possible way of increasing the robustness to noise is extending the training dataset with these noisy character images to train the character recognizer further. But this approach is costly, and it's hard to expose all possible noises to recognizers in training. 

Intrinsically, a character is constructed by a sequence of strokes. This property can be used to design a more robust character recognition method. We can extract the strokes first and reconstruct the characters, then further conduct the recognition.

\begin{figure}[t]%
    \centering
    \includegraphics[width=0.5\textwidth]{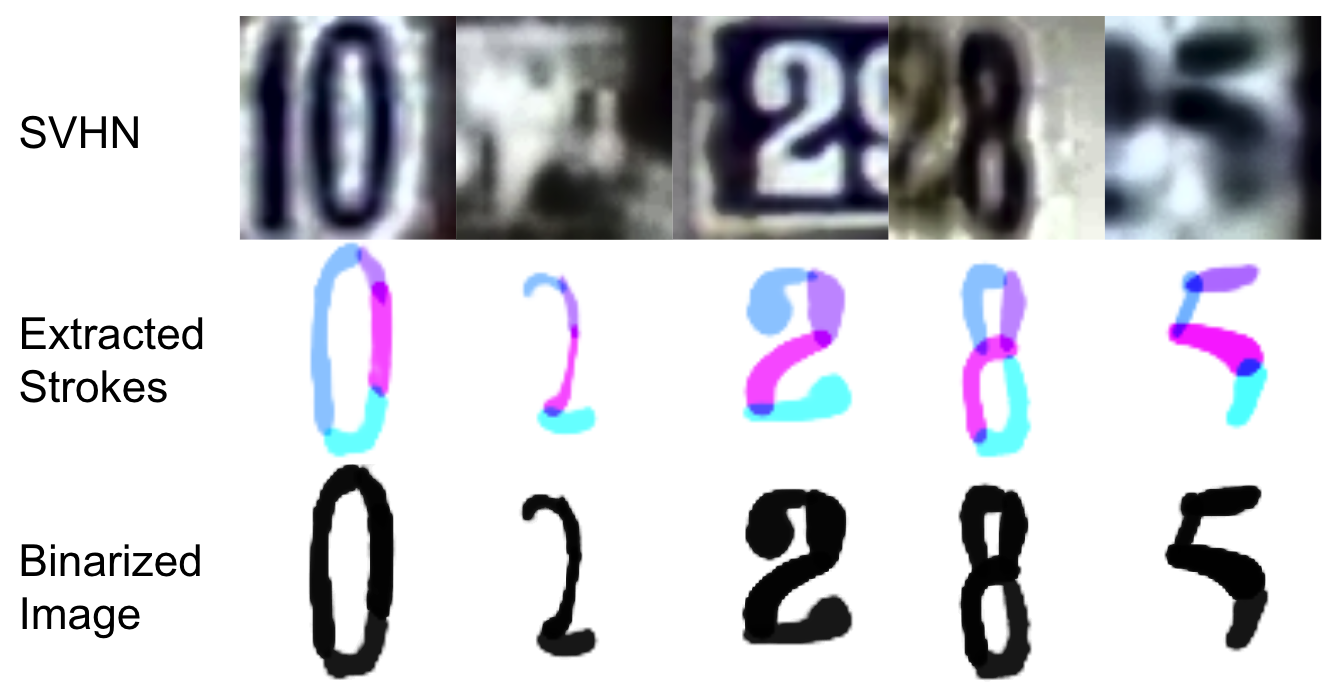}
    \caption{Extracted strokes and reconstructed images for images of SVHN. We only train our model on our synthetic SVHN-like data.}
\end{figure}

\begin{figure}[t]%
    \centering
    \includegraphics[width=0.5\textwidth]{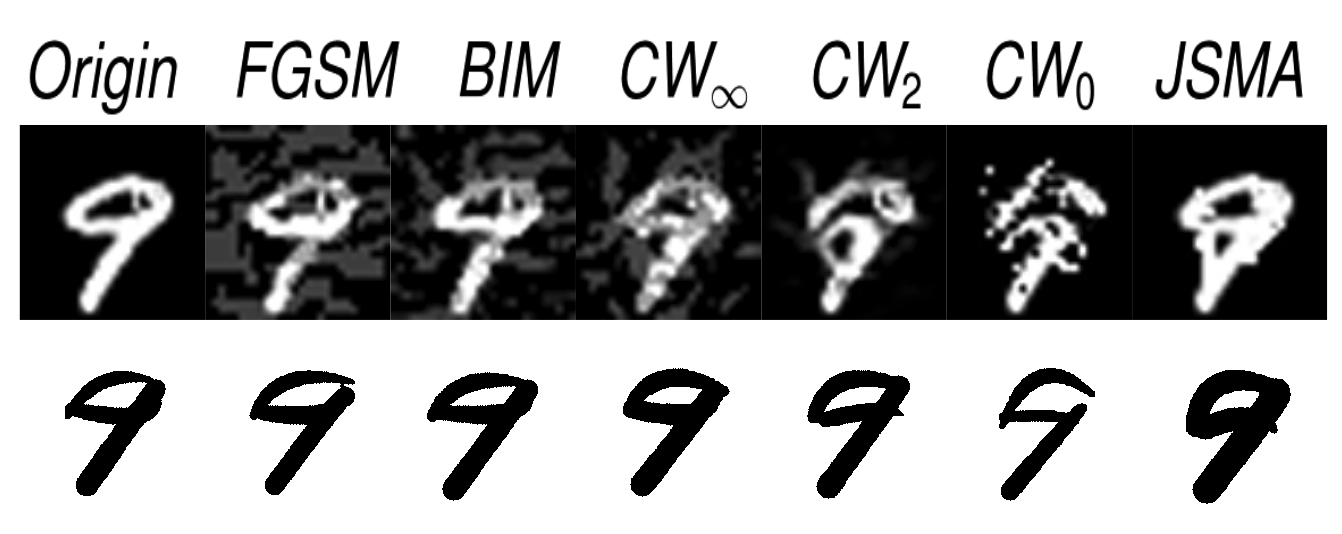}
    \caption{Reconstruction results for adversarial examples of MNIST.}
\end{figure}

\begin{figure*}[t]%
    \centering \includegraphics[width=0.8\textwidth]{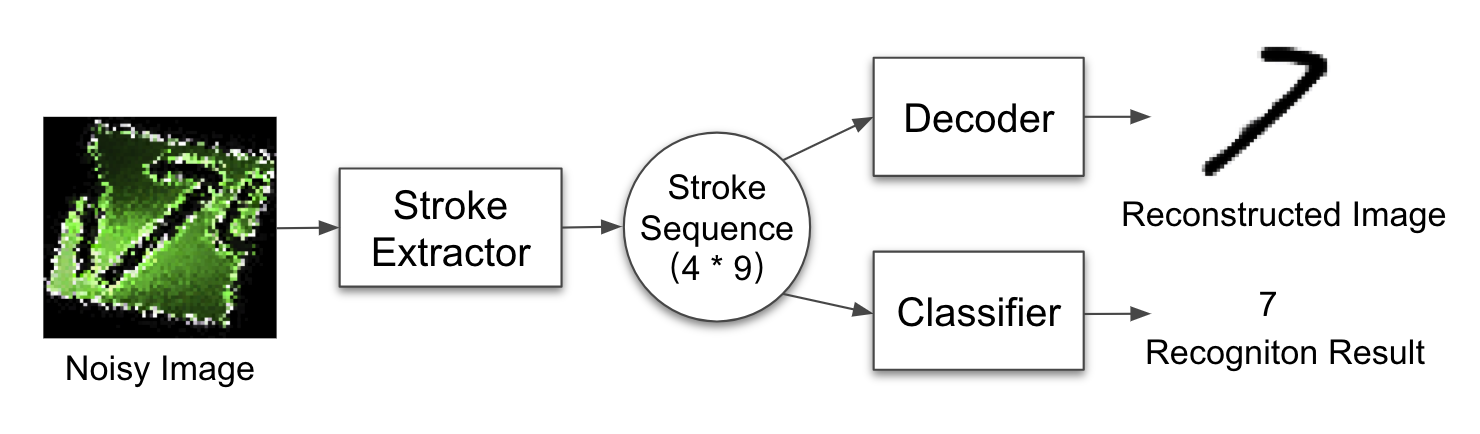}
    \caption{Pipeline of our SCR method. We use four curves to represent a character; each curve has nine parameters.}
    \label{fig:pipeline}
\end{figure*}
In this paper, we propose an encoder-decoder structure based on stroke representation which can deconstruct a character image and encode it into parameters of some strokes, then decode these parameters to an image. The weighted quadratic Bézier curve(WQBC) is used to simulate the stroke, which is controlled by three weighted points. We pre-train a neural network as the decoder that can embed the WQBC on an image based on the parameters of it. Then we freeze the decoder and train another neural network as an end-to-end stroke extractor, which can be integrated with character recognizers. The extracted strokes can be used in two ways: (1) reconstructing a clean character image for static image-based character recognizer; (2) taking as the input for stroke sequence based character recognizer. The pipeline of our SCR method is illustrated in Fig.~\ref{fig:pipeline}.

We take the $L_2$ distance the reconstructed character and the ground truth as the supervised loss and don't expose the strokes sequence of each character to the decoder. Despite all this, we find the decoder still can automatically learn to deconstruct a character into some strokes. And in experiments, we show that the decoder can cope with both the Arabic numerals and English letters well.

Our contributions are as follows:
\begin{itemize}  
\item We propose to encode character into the meaningful representation to improve the reconstruction ability and robustness of encoder-decoder structure and achieve better reconstruction effect.

\item We train a differentiable curve parameter decoder.

\item We design an image augmentation process for the training of our stroke extractor. Finally, our SCR method can deal with various image noise and can be well generalized in real scenes. We don't use images of SVHN for supervised training or transfer training but achieve a general reconstruction success rate of 89\%.

\item Our stroke extractor for scene character images can be integrated with character recognizer to improve the defense against adversarial attacks.
\end{itemize}  

%% file: related.tex
\section{2.Related Work}
\subsection{Character Reconstruction}
Some work attempts to improve the performance of character recognition with background image elimination. \cite{shen2015improving} However, these implementations are usually based on traditional methods of contrast adjustment, sharpening, binarization, etc., and require many manual designs. Among them, only the deblurring of the document is perfect\cite{chen2011effective}.

\subsection{Stroke Extractor}
Some works focus on the methods of reinforcement learning(RL) that make the agent have the ability to extract strokes from images. The work of 2013 studied the automatic real-time generation of simple strokes~\cite{xie2013artist}. DeepMind proposed an approach to train agent with reinforced adversarial learning to master synthesizing programs for images and use strokes to reconstruct images~\cite{ganin2018synthesizing}. 

\subsection{Adversarial Attack and Defense}
Robustness is essential to guarantee the security of application of machine intelligence. Unfortunately, many works have confirmed that deep neural networks are susceptible to small perturbations of the input vector~\cite{goodfellow2014explaining}. Current defense method against adversarial examples follows about three approaches: (1) Training to distinguish common and adversarial examples. (2) Training with adversarial examples to improve robustness. (3) Preprocessing input data or some other methods to make it difficult to attack target classifier~\cite{meng2017magnet}. Both (1) and (2) require many adversarial examples to train the models, but adversarial examples could be generated by unpredictable methods. % This pose potential security threats to artificial intelligence systems in practice.

%% file: env.tex
\section{3.SCR}
\subsection{Problem Definition}
Given a distorted character image $I_{dis}$, our goal is to reconstruct a clean character image $I_{rec}$ that is close to the ground truth character $I_{gt}$. We do not intend to crop the character from $I_{dis}$ at a time, because this is quite hard for machines, especially when $I_{dis}$ is with severely distortions. Instead, we deconstruct the character into parameters of strokes and draw strokes on the canvas. 

\subsection{Stroke Representation}
We use the \textbf{weighted quadratic Bézier curve} (WQBC) to simulate a stroke. The WQBC is determined by three weighted points $P_0$, $P_1$ and $P_2$, either of which is denoted as $(x,y,w)$, where $x$ and $y$ denote the coordinate and $w$ denotes the radius. Then a WQBC is generated as follows:
\begin{eqnarray*}
B(t) = (1-t)^2P_0 + 2(1-t)tP_1 + t^2P_2, 0\leq t \leq 1
\end{eqnarray*}
We have also tried other curves to simulate strokes. But we find WQBC is simple and flexible enough to simulate the most strokes of common characters. For the WQBC drawing program please refer to \textbf{Algorithm} ~\ref{alg:1}. The curve is drawn on a higher resolution canvas $(256 \times 256)$ to eliminate aliasing.

\begin{algorithm}[h]
\caption{\small Algorithm for drawing a WQBC} 
\raggedright
\hspace*{0.00in} {\bf Input:} Parameters of a WQBC, including $x_0, y_0, w_0, x_1, y_1, w_1, x_2, y_2, w_2$, they range from 0 to 1.\\
\hspace*{0.0in} {\bf Output:} Embedded image, its size is $64 \times 64$.\\
\hspace*{0.00in} {\bf Initialization:} Create a canvas C, which is a full 0 array of $256 \times 256$.\\
\begin{algorithmic}[1]
\raggedright
\State map $x_0, y_0, x_1, y_1, x_2, y_2$ to $[0,255]$, map $w_0, w_1, w_2$ to $[2, 32]$.
\State $i=0$
\While{$i<T$} 
    \State $t = i * (1 / T)$
    \State x = $(1-t)^2 * x_0 + 2 * t * (1-t) * x_1 + t^2 * x_2$
    \State y = $(1-t)^2 * y_0 + 2 * t * (1-t) * y_1 + t^2 * y_2$
    \State w = $(1-t)^2 * w_0 + 2 * t * (1-t) * w_1 + t^2 * w_2$
    \State draw a circle at $(x,y)$ of C, its radius is $w$. 
    \State $i=i+1$
\EndWhile
\State I = resize C to $64 \times 64$
\State \Return I
\end{algorithmic}
\label{alg:1}
\end{algorithm}

\subsection{Stroke Extractor}
The stroke extractor can deconstruct the character image into parameters of several WQBCs. We find that VGGNet-like architecture has well fitting and generalization capabilities on this task. We use a small size vggnet with halved number of channels as the stroke extractor.

\subsection{Differentiable Decoder}
Embedding the curve based on its parameters requires a method of calculating the distance from a point to a quadratic line segment. Performing such calculations with a differentiable operation is cumbersome, especially since the image is a collection of discrete pixels with size and value limits. When we want to try other different curve representations, we need to re-complete the derivation. One solution is to train an agent with reinforcement learning to implicitly build such a decoding model, but continuous space control is a problem in reinforcement learning. We make some attempts in our earlier version but don't achieve good enough effect.

We train a neural network that combines fully connected layers and upsampling layers. We use supervised learning methods to make the network learn to generate curves like programs. Parameters are uniformly sampled from the range during training. The network structure is shown in the appendix. It can embed WQBCs on a $64 \times 64$ image, and the inferred curve is very subtly different from the WQBC drawn by the program. 
\begin{figure}[htb]%
\centering
\includegraphics[width=0.5\textwidth]{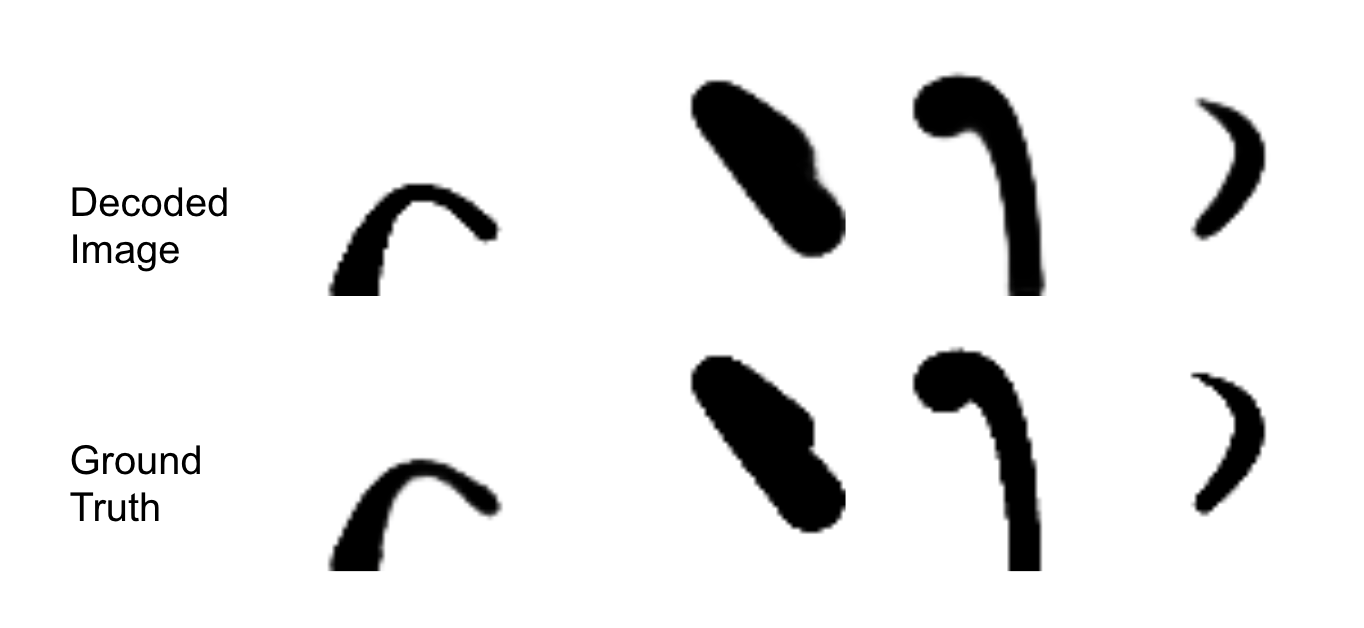}
    \caption{Decoded WQBC and ground truth. }
    \label{fig:Bézier}
\end{figure}
\subsection{Training}

The implementation of our training is based on the PyTorch v0.4 library~\cite{paszke2017automatic}. We use stochastic gradient descent(SGD) with momentum~\cite{sutskever2013importance} for training the neural network with a learning rate of $3\times10^{-2}$. For every 1/3 of training, the learning rate is reduced to 1/10.

%% file: exper.tex
\section{4. Experiment}
\subsection{Datasets and Noisy Characters Generation}
\subsubsection{MNIST}
MNIST~\cite{lecun1998mnist} contains 70,000 examples of hand-written digits, of which 60,000 are training data and 10,000 are testing data. Each example is a grayscale image with a resolution of $28\times28$ pixels. 

\subsubsection{SVHN} 
SVHN~\cite{netzer2011reading} is a real-world image dataset including over 600,000 digit images. The ``Cropped Digits" set is similar to MNIST, and each sample is a color image with a resolution of $32\times32$ pixels.

\subsubsection{Synthetic SVHN}
We collected 250 fonts to generate 2,500 Arabic numerals and 2,500 images from the MNIST training set, using these glyphs to synthesize an SVHN-like data set. We stitch together three digital pictures horizontally each time as ground truth. We randomly set the color of the characters and background then add various types of distortions and noise to ground truth to get the distorted image. We synthesize 200,000 images for training.

% We convert them into grayscale for training and testing.

%\subsubsection{Synthetic English Letters}
%We choose 4 English letters fonts to generate a dataset containing 104 ($4\times26$) images. Each image shows a printed lowercase English letter and with a resolution of $64\times64$ pixels.

\subsubsection{Noisy Characters Generation} 
We mainly consider following distortions and use an open source tool named imgaug\footnote{\url{https://github.com/aleju/imgaug/}} to achieve most image augments. A portion of the augmentation is performed at random each time. Among them, the augmentations involving displacement and deformation are performed on both ground truth and distorted image. Each pixel value of the image is normalized into the range of $[0,1]$ at first. 

\begin{itemize}
\item \textbf{Random Rotation} The rotation angle is randomly selected from the range of $[-15^\circ, 15^\circ]$. 

\item \textbf{Crop and Pad}
The width of the cropped image is between 60\% and 90 \% of the original image. 

\item \textbf{Gaussian Noise}  
We add a Gaussian noise $(\sigma \in (0, 0.05))$. 

\item \textbf{Blur} We choose one of the Gaussian blur$(\sigma \in [0, 3])$, the median blur$(r \in [3, 9])$, and the average blur$(r \in [2, 7])$ to act on the image.

\item \textbf{Sharpen and Emboss}
We use image sharpen $(alpha \in (0, 1.0), lightness \in (0.75, 1.5))$ and Emboss$(alpha \in (0, 1.0), strength \in (0, 2.0))$ operation.

\item \textbf{Pepper \& Salt Noise} The Signal-to-noise ratio is set to s $(s \in [0.7, 0.97])$.

\item \textbf{Other distortions}
We use perspective transform$(scale \in (0.01, 0.1))$ and piecewise affine $(scale \in (0.01, 0.05))$ operations provided by imgaug.

\end{itemize}

\subsection{Defense against Adversarial Attacks on MNIST}

Many works show that the written-digit recognizers are easily fooled by existing adversarial attack methods. 

\begin{figure}[ht]%    
    \centering       \includegraphics[width=0.5\textwidth]{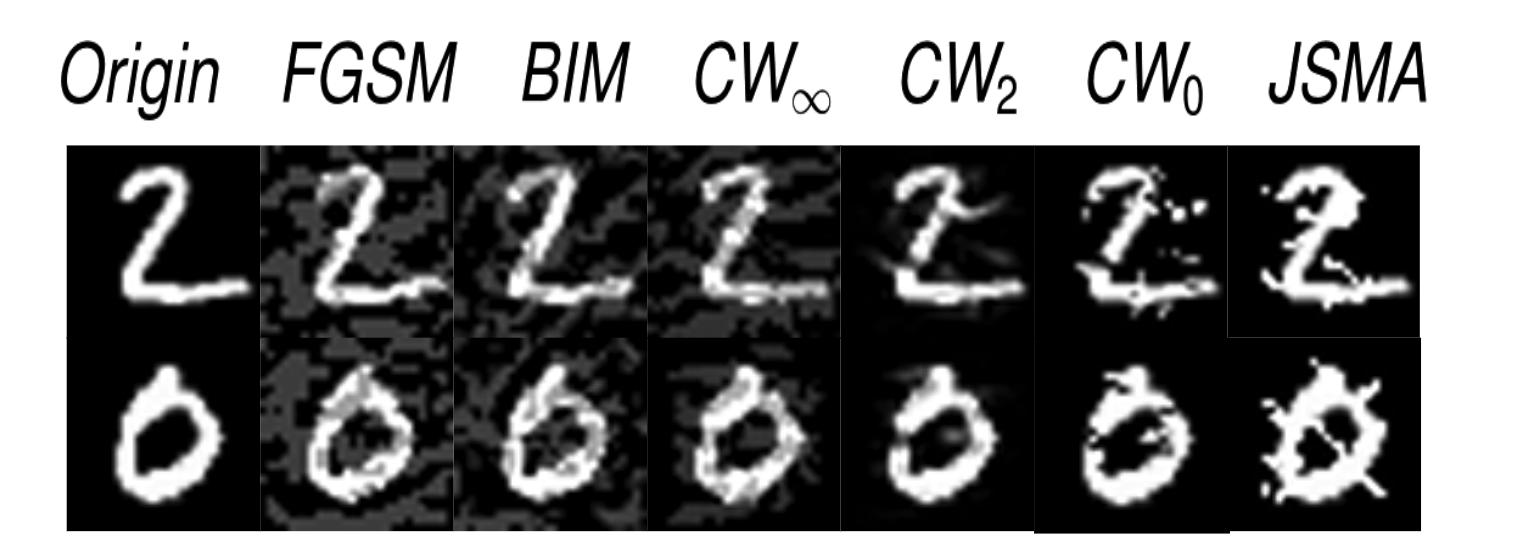}
    \caption{Adversarial examples of MNIST}
    \label{Adversarial_examples}
\end{figure}

Based on the MNIST dataset, we make a comparison to an existing defense method, feature squeezing~\cite{xu2017feature}. As in their paper, we take some adversarial attack methods in our experiments and show some adversary examples in Fig~\ref{Adversarial_examples}.

We take the training set (60,000 images) of MNIST to train the agent. These images are considered as the ground truth. In training, noisy character images are generated following the noisy character generation procedure. 

\begin{table*}[htb]
\centering
\begin{tabular}{|c|c|c|c|c|c|c|c|c|c|c|c|}
\hline
\multirow{3}{*}{\begin{tabular}[c]{@{}c@{}}MNIST\\ Experiments\end{tabular}} & \multicolumn{4}{c|}{$L_\infty$ Attacks}                                                & \multicolumn{2}{c|}{$L_2$ Attacks} & \multicolumn{4}{c|}{$L_0$ Attacks}                               & \multirow{3}{*}{All Attacks} \\ \cline{2-11}
                                                                             & \multirow{2}{*}{FGSM} & \multirow{2}{*}{BIM} & \multicolumn{2}{c|}{$CW_\infty$}        & \multicolumn{2}{c|}{$CW_2$}        & \multicolumn{2}{c|}{$CW_0$}      & \multicolumn{2}{c|}{JSMA}     &                              \\ \cline{4-11}
                                                                             &                       &                      & Next           & LL             & Next           & LL             & Next          & LL            & Next          & LL            &                              \\ \hline
No defense                                                                        & 54\%                  & 9\%                  & 0\%            & 0\%            & 0\%            & 0\%            & 0\%           & 0\%           & 27\%          & 40\%          & 13.00\%                      \\ \hline
Bit Depth (1-bit)                                                                    & 92\%         & \textbf{87\%}                 & \textbf{100\%} & \textbf{100\%} & 83\%           & 66\%           & 0\%           & 0\%           & 50\%          & 49\%          & 62.70\%                      \\ \hline
Median Smoothing (3x3)                                                             & 59\%                  & 14\%                 & 43\%           & 46\%           & 51\%           & 53\%           & 67\%          & 59\%          & 82\%          & 79\%          & 55.30\%                      \\ \hline
SCR (ours)                                                  & \textbf{94}\%                  & 86\%                 & 97\%           & 92\%           & \textbf{94\%}  & \textbf{83\%}  & \textbf{86\%} & \textbf{79\%} & \textbf{91\%} & \textbf{84\%} & \textbf{88.60\%}             \\ \hline

Human Performance                                                  & 97\%                  & 88\%                 & 97\%           & 97\%           & 94\%  & 78\%  & 83\% & 76\% & 89\% & 90\% & 88.90\%             \\ \hline
\end{tabular}
\caption{Model accuracy against adversarial attack}
\label{MNIST-adversary}
\end{table*}

In Table~\ref{MNIST-adversary}, we show the defense results of our method and feature squeezing, including reducing the color bit depth of each pixel and spatial smoothing. It's been demonstrated that our method achieves the best defense results for most adversarial attacks. And the overall performance of our method is significantly better than other methods.

\subsection{Experiments on SVHN}
\begin{figure}[ht]%
    \centering    \includegraphics[width=0.5\textwidth]{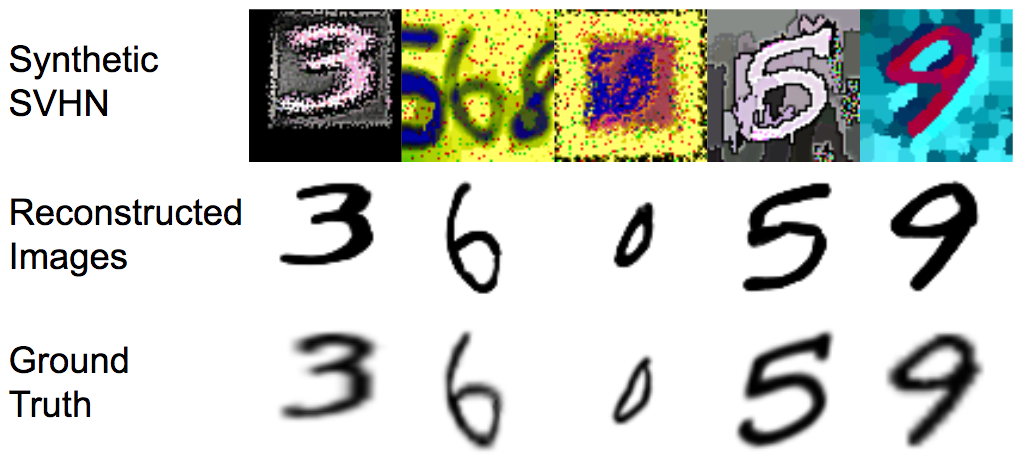}
    \caption{Reconstructed images for images of synthetic SVHN.}
   \label{Fake_SVHN}
\end{figure}
The character images of SVHN are cropped from real-scene street view images, and they are with rich noises in nature. Since it's hard for us to obtain the ground truth images (clean and without noise) for SVHN, we take a special strategy to train the stroke inference agent. We synthesize noisy character images based on MNIST and character images generated by system fonts as shown in Fig~\ref{Fake_SVHN}. Then take them to train our networks. When testing, we use the trained agent to infer stroke for SVHN. The agent still achieves a higher than 89\% recognition rate on SVHN. 
\subsection{Experiments on synthetic English letters}

Training on synthetic English letters is similar to the training on MNIST. We use the same noisy character generation procedure to generate noisy character images in training and testing.

\begin{figure}[ht]%
    \centering
    \includegraphics[width=0.5\textwidth]{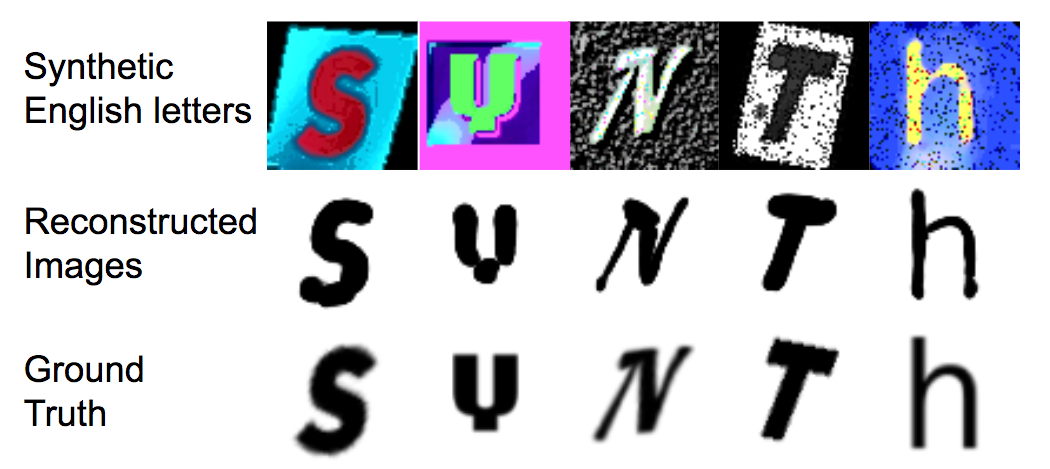}
    \caption{Reconstructed images for images of synthetic English letters}
    \label{English}
\end{figure}

Some stroke inference results are shown in Fig~\ref{English}.

%% file: discussion.tex
\section{5. Discussion}

The core part proposed SCR method is the stroke inference module, which is implemented as a stroke extractor and a differentiable decoder. The experiments show that the stroke extractor robust to the noise on characters. And this SCR method shows excellent ability in defending adversarial attacks of the hand-written digits. 

%In our future work, we will pay attention to the following aspects. 
%\begin{itemize}
%\item Improve our training algorithm and try some other expressive types of strokes to deal with complex characters such as Chinese, Arabic or calligraphy.
%\item Extend current method from single-character recognition to multi-character recognition. 
%\item Learn a general agent for multiple characters. It's usually hard to learn an agent for some specific character types, due to the difficulty in obtaining the ground truth characters. We hope to train an agent that can infer strokes for different characters, even without having access to them during training. 
%\end{itemize}

%% file: appendix.tex
\section{Appendix}
\subsection{Network structure}
The network structure diagram is shown in Figure~\ref{fig:encoder} and Figure~\ref{fig:decoder}, where \texttt{FC} refers to a fully-connected layer, \texttt{Conv} is a $3 \times 3$ convolution layer. The shape representation format of the output of the convolutional layer is [H, W, C]. All ReLU\cite{nair2010rectified} activations between the layers have been omitted for brevity. The activation function after stroke extractor and decoder is a sigmoid function in order to map the output into range [0,1].
\begin{figure}[ht]
\centering
\input{architecture/encoder.tex}
\caption{The architecture of the \textbf{stroke extractor} and \textbf{classifier}.}
\label{fig:encoder}
\end{figure}
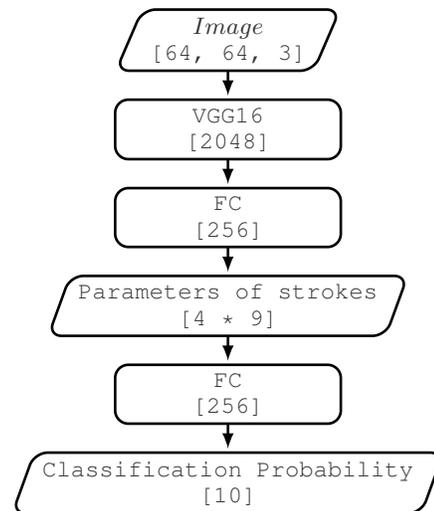
\begin{figure}[ht]
\centering
\input{architecture/decoder.tex}
\caption{The architecture of the \textbf{decoder}.}
\label{fig:decoder}
\end{figure}
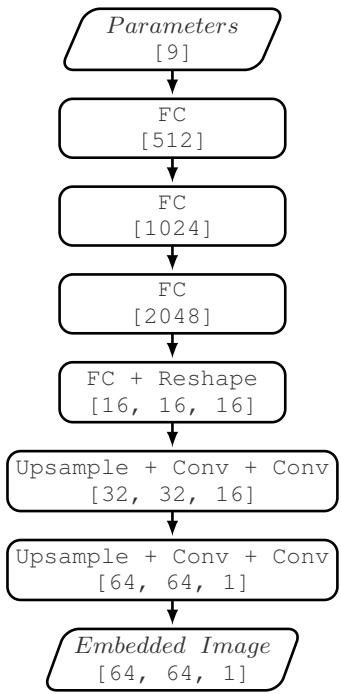

%% file: architecture/encoder.tex
\begin{tikzpicture}[
  scale=0.3,
  black!100,
  text=black!80,
  node distance=0.35cm and 0.7cm,
  % Base flowchart node.
  fnode/.style={
    align=center,
    % The shape.
    rectangle,minimum height=15pt,minimum width=3cm,rounded corners=5pt,
    % The rest.
    inner sep=3pt,
    line width=1pt,
    fill=pnodefill,draw=pnodedraw,
    font=\small\ttfamily},
  nonode/.style={
    align=center,
    font=\small\ttfamily},
  % Data node.
  dnode/.style={
    fnode,trapezium,trapezium left angle=70, trapezium right angle=110,trapezium stretches},
  % Input node.
  inode/.style={
    dnode,fill=inodefill,draw=inodedraw},
  downnode/.style={
    fnode,trapezium,trapezium left angle=110, trapezium right angle=110,trapezium stretches},
  % Process node.
  pnode/.style={
    fnode,fill=pnodefill,draw=pnodedraw},
  % Process node (variation).
  pnode2/.style={
    fnode,fill=pnodefill!50!white,draw=pnodedraw!50!white},
  % Output node.
  onode/.style={
    dnode,fill=onodefill,draw=onodedraw},
  smnode/.style={
    mnode,fill=mnodefill,draw=mnodedraw},
  flow/.style={
    -latex,shorten >=1pt,line width=1pt,line cap=round,rounded corners=2pt,draw=pnodedraw,draw==\#1},
  flow2/.style={
    flow,draw=pnodedraw!50!white}]
  \newcommand{\mytrap}[3]{%
    \node (#1) [anchor=center,#2] {\phantom{$ State $}\\\phantom{[64,64,1]}}; \node[anchor=center,align=center,font=\small\ttfamily] at (#1.center) {#3};
  }

    \mytrap{input}{dnode}{$ Image $\\{[64, 64, 3]}};
  \node (fc0) [pnode, below= of input] {VGG16\\{[2048]}};
  \path (input) edge[flow] (fc0);
  \node (fc1) [pnode, below= of fc0] {FC\\{[256]}};
  \path (fc0) edge[flow] (fc1);
  \node (fc2) [dnode, below= of fc1] {Parameters of strokes\\{[4 * 9]}};
  \path (fc1) edge[flow] (fc2);
  \node (fc3) [pnode, below= of fc2] {FC\\ {[256]}};
  \path (fc2) edge[flow] (fc3);  
  	\node (output) [dnode, below= of fc3] {Classification Probability \\{[10]}};
  \path (fc3) edge[flow] (output);
  \end{tikzpicture}

%% file: architecture/decoder.tex
\begin{tikzpicture}[
  scale=0.3,
  black!100,
  text=black!80,
  node distance=0.35cm and 0.7cm,
  % Base flowchart node.
  fnode/.style={
    align=center,
    % The shape.
    rectangle,minimum height=15pt,minimum width=3cm,rounded corners=5pt,
    % The rest.
    inner sep=3pt,
    line width=1pt,
    fill=pnodefill,draw=pnodedraw,
    font=\small\ttfamily},
  nonode/.style={
    align=center,
    font=\small\ttfamily},
  % Data node.
  dnode/.style={
    fnode,trapezium,trapezium left angle=70, trapezium right angle=110,trapezium stretches},
  % Input node.
  inode/.style={
    dnode,fill=inodefill,draw=inodedraw},
  downnode/.style={
    fnode,trapezium,trapezium left angle=110, trapezium right angle=110,trapezium stretches},
  % Process node.
  pnode/.style={
    fnode,fill=pnodefill,draw=pnodedraw},
  % Process node (variation).
  pnode2/.style={
    fnode,fill=pnodefill!50!white,draw=pnodedraw!50!white},
  % Output node.
  onode/.style={
    dnode,fill=onodefill,draw=onodedraw},
  smnode/.style={
    mnode,fill=mnodefill,draw=mnodedraw},
  flow/.style={
    -latex,shorten >=1pt,line width=1pt,line cap=round,rounded corners=2pt,draw=pnodedraw,draw==\#1},
  flow2/.style={
    flow,draw=pnodedraw!50!white}]
  \newcommand{\mytrap}[3]{%
    \node (#1) [anchor=center,#2] {\phantom{$ State $}\\\phantom{[64,64,1]}}; \node[anchor=center,align=center,font=\small\ttfamily] at (#1.center) {#3};
  }

    \mytrap{input}{dnode}{$ Parameters $\\{[9]}};
  \node (fc0) [pnode, below= of input] {FC\\{[512]}};
  \path (input) edge[flow] (fc0);
  \node (fc1) [pnode, below= of fc0] {FC\\{[1024]}};
  \path (fc0) edge[flow] (fc1);
  \node (fc2) [pnode, below= of fc1] {FC\\{[2048]}};
  \path (fc1) edge[flow] (fc2);
  \node (fc3) [pnode, below= of fc2] {FC + Reshape \\ {[16, 16, 16]}};
  \path (fc2) edge[flow] (fc3);  
  \node (conv0) [pnode, below= of fc3] {Upsample + Conv + Conv \\ {[32, 32, 16]}};
  \path (fc3) edge[flow] (conv0);  
  \node (conv1) [pnode, below= of conv0] {Upsample + Conv + Conv \\ {[64, 64, 1]}};
  \path (conv0) edge[flow] (conv1);  
  	\node (output) [dnode, below= of conv1] {$ Embedded \ Image $\\{[64, 64, 1]}};
  \path (conv1) edge[flow] (output);
  \end{tikzpicture}